# Position Prediction of Ball and Fuzzy Controller for Shooting Action in A Soccer Robot System

GyongIl Ryang, MyongSong Choe, YongChol Sin

Faculty of Electronics & Automation, **Kim Il Sung** University, Pyongyang

Democratic People's Republic of Korea

**Abstract:** The robot soccer game based complex motion control has been widely studied for the moving object capture and shooting. A position prediction algorithm based on global vision is introduced in order to improve the accuracy and robustness of the vision system for tracking moving objects, including a *Kalmanfiter*, a dynamic window and an obstacle avoidance strategy. This paper deals with the positon prediction for moving ball by using *Kalmanfiter* and the *Fuzzy Controller* for shooting action in a dynamic environment.

## 1. Introduction

The soccer robot system consists of 3 soccer robots per a team and Robot football system includesmicrorobotics, artificial intelligence, multi agent real time control and dynamic vision sensing in a unique way. It has to estimate ball position by processing an image from a CCD camera, and it make a choice of next action in a Decision-Making system, and then it transmit motion speed signal of soccer robots in a Wireless Communication system.

The following picture shows the structure of soccer robot system for Mirosot((Micro Robot Soccer Tournament)game. (Figure. 1)

There are several basic skills such as moving, shooting, dribbling and intercepting in a robot soccer game and the most important skill is the *Shooting*. By a good motion controller,it has improved a high accuracy of action.

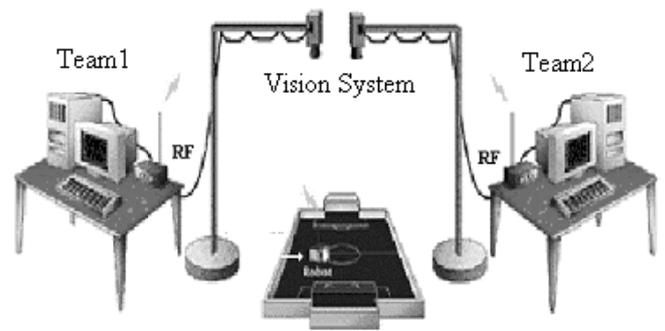

Figure 1. Structure of Soccer robot system

The kinematic model of soccer robot is expressed as follows.[2]

$$\begin{cases} \dot{x} = v \cdot \cos\theta \\ \dot{y} = v \cdot \sin\theta \\ \dot{\theta} = \omega \end{cases} \quad (1)$$

The posture of soccer robot can be represented as $(x, y, \theta)$, where $(x, y)$ is the position of central point and $\theta$ is a head angle between X-axis and face of soccer robot, and $v, \omega$ are linear and angular velocities respectively.

To design Motion controller is to determine the left and right speeds of robot so that it moves a desired posture $(x_d, y_d, \theta_d)$ from any initial one. [2-3]

A Proportional Control law[4] is one that calculate the control force in propotion to error between a desired posture and a current one, and it is expressed as follows.

$$v_l = k_d \cdot d_e - k_\theta \cdot \theta_e$$
$$v_r = k_d \cdot d_e + k_\theta \cdot \theta_e \qquad (2)$$

The proportional controller have acheived zero error between target and current position, but not a desired head angle at the point. So control force has to be recalculated for turning robot to a desired one.

A Fuzzy logic control law[5] is one that sub-controllers are the proportional ones and a entire control force is calculated by fuzzy logic combination of these sub-controllers.

However, the high game performance depends on the shooting action, and it is important to predict the ball position in real-time and move soccer robot to the predicted point correctly and quickly.

Here, we formulate the following two problems.

First, the predictionof the ball postion in real time by using Kalman filter. Second, the design of the fuzzy controller to be movedsoccer robot to the correct position with keeping in the head angle of soccer robot smoothly

## 2. Prediction of ball position and Design of Fuzzy controller

1) Prediction of ball position

In Mirosot(Micro Robot Soccer Tournament) game, the ball moves to any position continually without stopping, so it is important to predict the ball position in advance. The vision,however, is often used as the input sensor of the environment and several approaches have been proposed for positon estimation. But it is difficult to kick the ball exactly without prediction of ball position, because the speed of ball is fast in the game. As the image processing is more time-consuming, it is important to reduce the size of image to be processed for predicting the position in real time.

We have introduced a Dynamic Window based object extraction as shown inFigure 2.

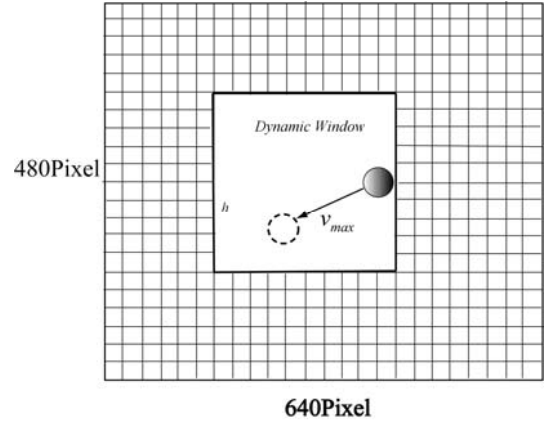

Figure 2. A Dynamic Window

The size of the square is defined as follows:

$$h = 2 \cdot r_{ball} + 2 \cdot v_{ball}^{max} \cdot T \quad (3)$$

$r_{ball}$ : the size of ball in image(Pixel),

$v_{ball}^{max}$ : the max speed of ball in the image(Pixel/s)

$T$ : sample time of the system (s)

[Algorithm for Position Estimation]

**Step 1**. Image capture: Image are captured in RGB in a $620 \times 480$ resolution

**Step 2.** Choosing the dynamic window according to the position in last snap.

**Step 3.** Segment the dynamic window into small square, the size of block is decided by the ball size in the image(usually $r_{ball}/4$) choosing the center of the square as a seed.

**Step 3.** Recursive algorithm to expand the area. Calculating the center of the region and filtering of the area including outside of the ball.

**Step 4.** Image Map to Global World

The least square method is adopted to fit the map

$$\begin{cases} x = a_0 + a_1 u + a_2 v + a_3 u \cdot v \\ y = b_0 + b_1 u + b_2 v + b_3 u \cdot v \end{cases} \quad (2)$$

, where $(x, y)$ is the ball position in the global world, and $(u, v)$ is the ball position in image, and $a_i, b_i, i = 0, \cdots, 3$ are coefficients of approximation function.

The Kalman filter can be described by the following equation to make prediction the ball position at next time.

$$\bar{\dot{x}}_n = \hat{\dot{x}}_{n-1} + h(z_{n-1} - \hat{x}_{n-1})/T \quad (5)$$

$$\bar{x}_n = \hat{x}_{n-1} + g(z_{n-1} - \hat{x}_{n-1}) \quad (6)$$

$\bar{\dot{x}}_n$: predicted estimate of ball velocity at time $n$,

$\hat{\dot{x}}_{n-1}$: estimation of ball velocity at time $n-1$ and all preceding times,

$z_{n-1}$: sensor reading at time $n-1$,

$\bar{x}_n$: predicted estimate of ball position at time $n$,

$\hat{x}_{n-1}$: estimation of ball position at time $n-1$ and all preceding times,

$T$: sample time in vision system,

$h$ and $g$ are filter parameters.

2) Design of Fuzzy controller for kicking up

In soccer robot system, the Shooting action is very important and it depends how well motion controller acts so that soccer robot move to target correctly and quickly from any posture. By using the preceding controllers, especially propotional and pure fuzzy one, the robot moves only forward. Therefore, it has to recalculate the control law for turning and back movement near to the target in order to improve the accuracy of control. However, in the game, the robot have to start from any posture and move to the target as soon as quickly and correctly. For this problem, it contains not only forward movement controller, but also turning and backward direction one.

So, we have proposed the Fuzzy controller combining with forward, backward turning movements by using Fuzzy logic. The following picture shows the fuzzy input variables to be selected.

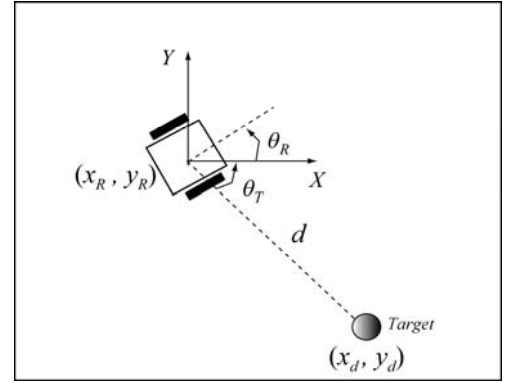

Figure 3. fuzzy input variables

① Forward Direction Controller

[Algorithm forForward Direction Movement]

**Step 1**: Determining the control gains: $k_d = 0.85$, $k_a = 0.12$

**Step 2**: Calculating postion and angle errors.

$$d = (x_d - x_R)^2 + (y_d - y_R)^2 \quad (7)$$

$$\theta_T = a\tan 2(y_d - y_R, x_d - x_R) \quad (8)$$

$$\theta_e = \theta_T - \theta_R \quad (9)$$

**Step 3**: Calculating the control law.

$$\begin{aligned} V_L^{forward} &= k_d \cdot d - k_a \cdot \theta_e \\ V_R^{forward} &= k_d \cdot d + k_a \cdot \theta_e \end{aligned} \quad (10)$$

, where $V_L^{forward}$, $V_R^{forward}$ are left and right speed of soccer robot respectively.

② Backward Direction Controller

[Algorithm forBackward Direction Movement]

**Step 1**: Determining the control gain: $k_d = 0.85$, $k_a = 0.12$

**Step 2**: Calculating postion and angle errors.

$$d = (x_d - x_R)^2 + (y_d - y_R)^2 \quad (11)$$

$$\theta_T = a\tan 2(y_d - y_R, x_d - x_R) \quad (12)$$

$$\theta_e = \theta_T - \theta_R \quad (13)$$

If $\theta_e$ is more than 180° ormore less -180°, then the angle is changeable to value between -180° and 180°.

**Step 3**: Calculating the control law.

$$V_L = k_d \cdot d + k_a \cdot \theta_e, \quad V_L^{Backward} = -V_L$$
$$V_R = k_d \cdot d - k_a \cdot \theta_e, \quad V_R^{Backward} = -V_R \quad (14)$$

③ Turning Controller

**[Algorithm for Turning Movement]**

**Step 1:** Determining the control gain: $k_a = 0.2$

**Step 2:** Calculating angle errors.

$$\theta_e = \theta_d - \theta_R \quad (15)$$

, where $\theta_d$ is the desired angle, and $\theta_R$ is the current one. If $\theta_e$ is more than 180° ormore less -180°, then the angle is changeable to value between -180° and 180°.

**Step 3**: Calculating the control law.

$$V_L^{Turn} = -k_a \cdot \theta_e$$
$$V_R^{Turn} = k_a \cdot \theta_e \quad (16)$$

The figure 4 shows the fuzzy input partition, and the fuzzy rules are as follows.

**[Fuzzy Rules]**

*Rule*1: If 《the position is *Near*》and 《$\theta_R$ is not 0》

Then $[V_L, V_R] = TurnControl(\theta_R, 0)$

… … …

*Rule* 26: If 《$\theta_R$ is 180》and 《$\theta_T$ is 180》

Then

$[V_L, V_R] = ForwardControl(x_R, y_R, \theta_R, x_d, y_d, \theta_T)$

**[Algorithm for a proposedFuzzy Control]**

**Step 1**: For Vision system, Estimation of the current position of ball soccer robots.

**Step 2:** Prediction of ball position and speed at next time by Kalman filter.

**Step 3:** Calculating distance and angle error between the predicted point and the current one.

**Step 4:** By using the proposed fuzzy controller, moving robot to the just back point of ball so that it can be convenient for kicking up.

**Step 5:** Determining the shooting direction taking consideration of the position of opposite goalkeeper robot.

**Step 6:** If the ball is very near to the robot, turning the head of robot to above shooting direction.

**Step 7:** Moving the robotto opposite goal along the straight linequicly for a very short time.

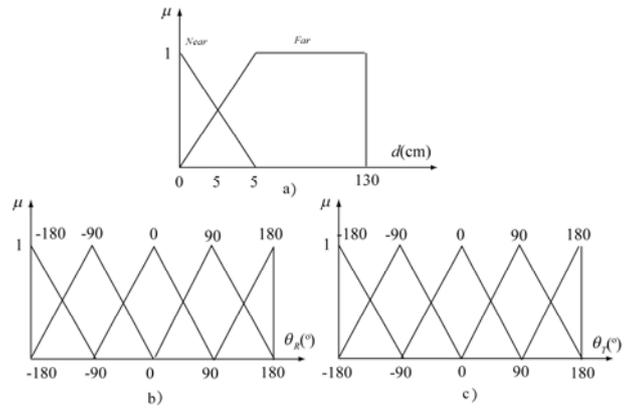

Figure 4. Fuzzy Input Partition

## 3. Experiments

To verify the proposed approach, we have the following experiment. Based on the calculation of the executive time of vision system, we have determined the sample time as 35ms and allocated 3 robots per a team, and it takes 5 minutes per a game.

The following table shows the comparision result of game between the proposed method and the preceding one. As you know, it records the more goals and it means that improve the performance of shooting action by using the proposed method and the Figure 5 shows the orbit of soccer robots .

table. Number of Goals

| N umber of round | 1 | 2 | 3 | 4 |
|---|---|---|---|---|
| Goals(preceding one) | 5(4) | 6(5) | 3(2) | 7(4) |

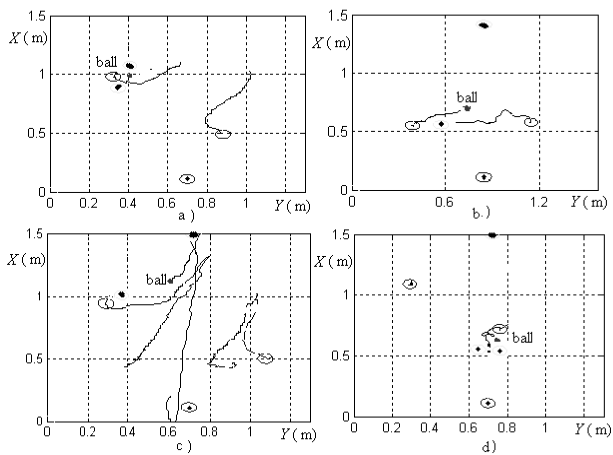

Figure 5. Prediction of ball and Shooting action

It shows that it have been predicted the ball position by using kalmanfilter correctly and the soccer robot have moved the position shooting the ball correctly. And it shows that under the inconvenient environment , it have moved rapidly by using the proposed fuzzy controller.

## Conclusions

In this paper, an approach for ball shooting using global vision system in a dynamic environment is presented.

First, we introduced a dynamic window to reduce the time of image processing and predicted the ball position which moved 8 robots continually in real-time.

Second, we proposed the fuzzy controller which can move robot to target starting from any posture smoothly, uses fuzzy combination with the basic motion controller such as forward, backward direction and turning one.

Lastly, the presented simulation shows good agreement of the proposed method with the real experiment.